\documentclass[sigconf]{acmart}
\usepackage{latexsym}
\usepackage{graphicx}
\usepackage{adjustbox}
\usepackage{multirow}
\usepackage{multicol}
\usepackage{dirtytalk}
\usepackage{subcaption}
\usepackage{amsmath}
\usepackage{arydshln}
\usepackage{listings}
\usepackage{color, colortbl}
\usepackage{tabularx}
\usepackage{amsfonts}
\usepackage{newtxmath}

\AtBeginDocument{%
  \providecommand\BibTeX{{%
    \normalfont B\kern-0.5em{\scshape i\kern-0.25em b}\kern-0.8em\TeX}}}

\copyrightyear{2018}
\copyrightyear{2024} 
\acmYear{2024} 
\setcopyright{acmlicensed}
\acmConference[CIKM '24] {Proceedings of the 33rd ACM International Conference on Information and Knowledge Management}{October 21--25, 2024}{Boise, ID, USA.}
\acmBooktitle{Proceedings of the 33rd ACM International Conference on Information and Knowledge Management (CIKM '24), October 21--25, 2024, Boise, ID, USA}

\acmISBN{979-8-4007-0436-9/24/10} 
\acmDOI{10.1145/3627673.3679562}
\begin{document}

\title{LLaVA-Chef: A Multi-modal Generative Model for Food Recipes
}


\author{Fnu Mohbat}
\affiliation{%
\institution{Rensselaer Polytechnic Institute}
 \city{Troy}
 \state {New York}
  \country{USA} 
  }
\email{mohbaf@rpi.edu}

\author{Mohammed J. Zaki}
\affiliation{%
  \institution{Rensselaer Polytechnic Institute}
  \city{Troy}
  \state {New York}
  \country{USA} 
  }
\email{zaki@cs.rpi.edu}




\begin{abstract}
In the rapidly evolving landscape of online recipe sharing within a globalized context, there has been a notable surge in research towards comprehending and generating food recipes. Recent advancements in large language models (LLMs) like GPT-2~\cite{radford2019language} and LLaVA~\cite{liu2023visual} have paved the way for Natural Language Processing (NLP) approaches to delve deeper into various facets of food-related tasks, encompassing ingredient recognition and comprehensive recipe generation. Despite impressive performance and multi-modal adaptability of LLMs, domain-specific training remains paramount for their effective application. This work evaluates existing LLMs for recipe generation and proposes LLaVA-Chef, a novel model trained on a curated dataset of diverse recipe prompts in a multi-stage approach. First, we refine the mapping of visual food image embeddings to the language space. Second, we adapt LLaVA to the food domain by fine-tuning it on relevant recipe data. Third, we utilize diverse prompts to enhance the model's recipe comprehension. Finally, we improve the linguistic quality of  generated recipes by penalizing the model with a custom loss function. LLaVA-Chef demonstrates impressive improvements over pretrained LLMs and prior works. A detailed qualitative analysis reveals that LLaVA-Chef generates more detailed recipes with precise ingredient mentions, compared to existing approaches.
  
\end{abstract}





\maketitle

\section{Introduction}

The significance of food for promoting well-being is growing, as a result understanding food recipes for healthy lifestyles has emerged as a critical research area. The recent growth of recipe data through online platforms and mobile apps has created a rich data resources, driving research efforts towards developing AI-powered solutions for food recognition, ingredient suggestion, and personalizing recipe, all while factoring in dietary restrictions, cultural preferences, and religious considerations ~\cite{papadopoulos2022cvpr, chhikara2023fire, salvador2019inverse, zhang2023understanding}. Despite substantial progress, generating recipes or cooking steps solely from food names, images, or ingredients remains a significant challenge. While the computer vision community has leveraged state-of-the-art deep learning techniques to extract ingredients from images, and NLP applications have facilitated recipe generation from food names or ingredients, the recent advances in multi-modal language-vision models offer a promising path towards crafting feasible real-world solutions by fusing visual and textual data.

\begin{figure*}[!ht]
    \centering
    \includegraphics[scale=0.46]{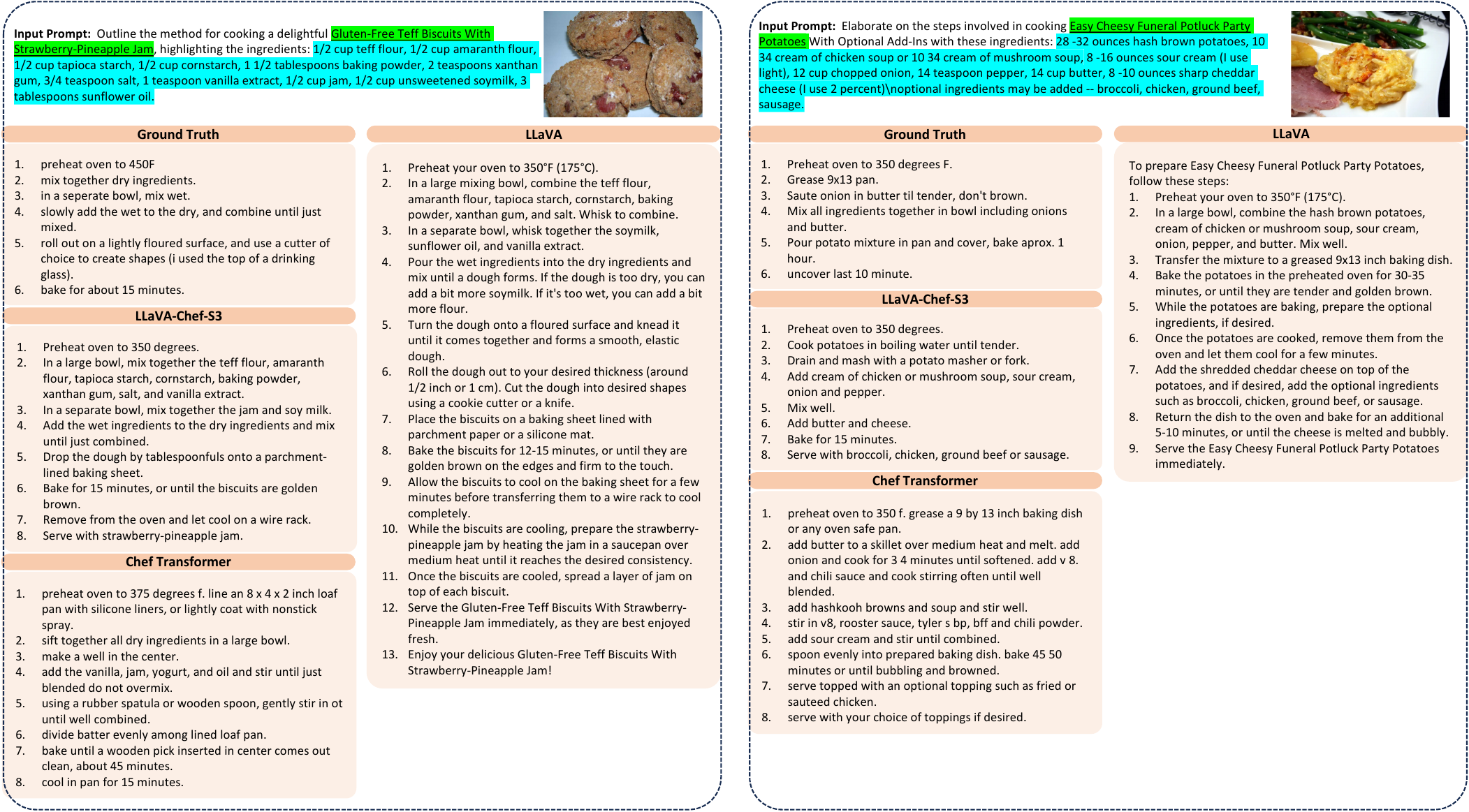}
    \caption{Sample recipes generated by LLaVA-Chef model, Chef-Transformer~\cite{farahani2023chef} (open source recipe generation model) and LLaVA~\cite{li2023llava} (best pretrained model).
    We can see issues of hallucination, repetitive test, and inaccuracies for previous models.}
    \label{fig:result_samples1}
    \Description[]{}
\end{figure*}

Large language models (LLMs)~\cite{raffel2020exploring, touvron2023llama, jiang2023mistral, mojan2023phi2} have demonstrated a remarkable ability to rapidly learn from vast amounts of text and even multi-modal data~\cite{liu2023visual,lai2023lisa, li2023blip}. For instance, by incorporating visual features extracted from pretrained vision-language models, several LLMs ~\cite{liu2023visual, li2023blip, awadalla2023openflamingo} have shown an enhanced ability to tackle vision-language tasks like image captioning, visual question answering, and visual reasoning. While these models excel in general applications, their expertise plummets when they encounter specialized domains due to insufficient domain-specific training~\cite{li2023llava,moor2023med}. This deficit often manifests in outputs riddled with hallucinations, inaccuracies, and repetitive text, as Figure~\ref{fig:result_samples1} demonstrates for food recipes generated by two models.


\begin{figure*}[ht]
    \centering
    \includegraphics[scale=0.58]{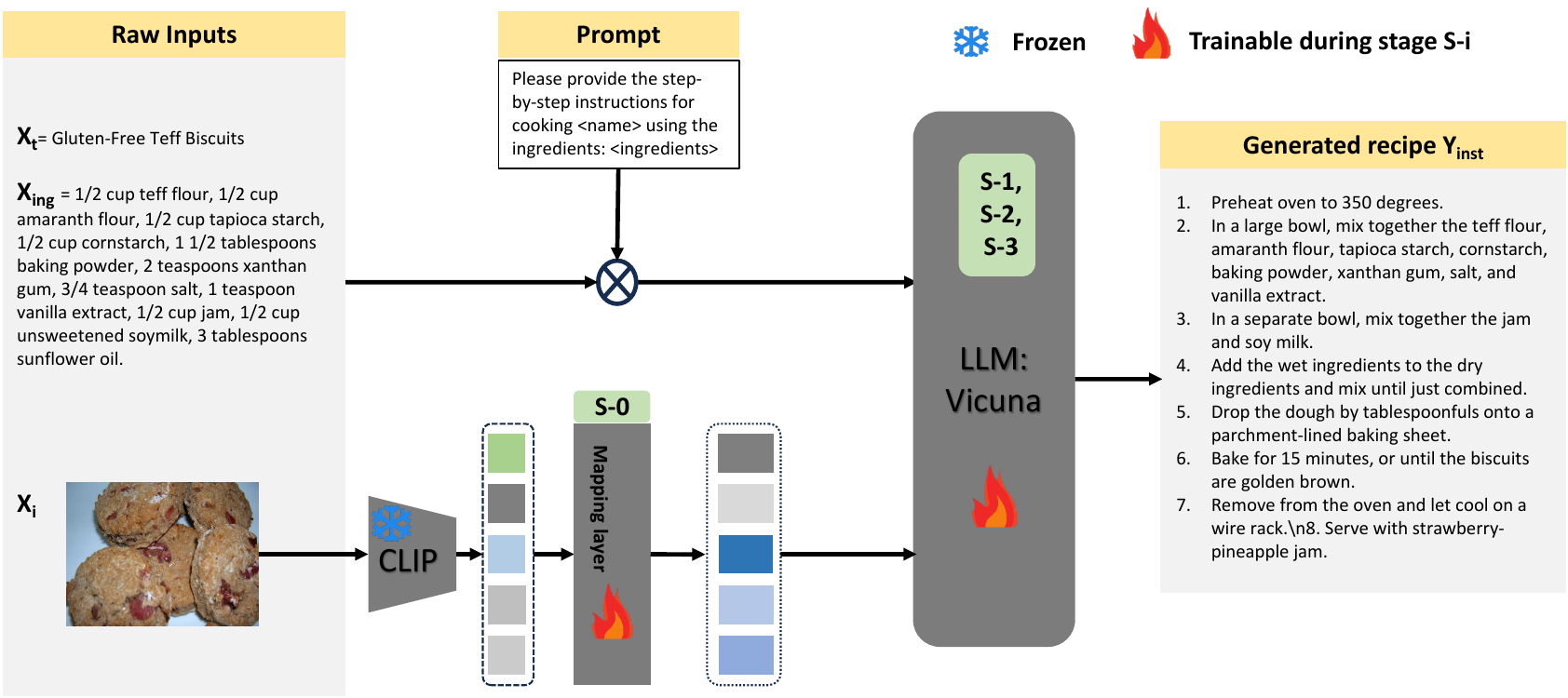}
    \caption{Architecture of LLaVA-Chef and different training stages (as shown in grey). The inputs to the model $X_t$, $X_{ing}$, and $X_i$ refer to the recipe {\normalfont {\bf t}itle, {\bf ing}redients and {\bf i}mage}, respectively. $Y_{inst}$ refers to the generated recipe {\normalfont {\bf inst}ructions} (which are compared with the ground truth instructions $X_{inst}$ for loss computation). In training Stage-0 ({\bf S-0}), the image to text mapping layer is fine-tuned. Whereas, in the rest of the training stages {\bf S-1, S-2, and S-3} the backbone LLM is fine-tuned. Given a recipe, we sample a prompt, then substitute <name> and <ingredients> with $X_t$ and $X_{ing}$. Visual features of the  image $X_i$ from CLIP are mapped in language space and concatenated with language embeddings before passing through the backbone LLM. The frozen and trainable symbols indicate which layers are fine-tuned (e.g., CLIP is frozen, whereas mapping layer and LLM are trainable.)}
    \label{fig:model1}
    \Description[]{}
\end{figure*}


Initial research focused on computer vision methods for food classification to ingredient detection ~\cite{chen2020study, he2021online, kaur2023deep,rodriguez2023dining}. 
Several researchers learned unique food embeddings using text-vision models~\cite{papadopoulos2022cvpr,rodriguez2023dining} while others generated food names using image captioning models ~\cite{chhikara2023fire}. Chef Transformer~\cite{farahani2023chef} takes a list of ingredients and generates recipes, whereas
~\cite{chhikara2023fire, tanejamonte, fatemi2023learning} predict ingredients from food images as an intermediate step towards recipe generation. One recent research ~\cite{yin2023foodlmm} fine-tuned the LISA~\cite{lai2023lisa} model for a variety of food tasks including food classification, recipe generation and segmentation. Despite various endeavors, none of the models have proven successful in generating effective recipes. Furthermore, most of these models lack robust evaluation or are not publicly available.

In this paper, we address the limitation of the existing methods by proposing LLaVA-Chef, a powerful multi-modal language and vision model for learning food recipes with the help of well curated and diverse set of prompts tailored towards training the model for food domain tasks. Our model extends the LLaVA~\cite{liu2023visual}, which consists of Vicuna~\cite{chiang2023vicuna} as a foundation LLM and CLIP~\cite{radford2021learning} as a visual encoder. The architecture of our model is shown in Figure ~\ref{fig:model1}. The model concatenates visual and textual embeddings, and inputs them to the backbone LLM to generate the desired output. Following ~\cite{li2023llava}, first we improve the cross-modal representation for food related images by fine-tuning the mapping. Then, the model is fine-tuned on unique prompts that reduce the hallucination and improve the quality of recipe text. 
In the following training stage, we improve the adaptability of the model for the food domain by introducing more than 100 unique prompts to generate different attributes of a recipe, i.e., title, ingredients and cooking instructions. Finally, we penalize the model with a novel scaling term based on text generation metrics, ultimately leading to improved performance. Thus, gradually involving the augmentation of prompt diversity and task complexity across multiple stages, our model systematically acquires proficiency in handling a wide array of food recipes. We evaluate our model on the Recipe1M dataset~\cite{salvador2017learning}, specifically on the test samples containing at least one image. 
Compared to pretrained LLMs, our model consistently achieves higher scores across most metrics. While other models could not get more than 0.1 CIDEr score, our model achieves a remarkable 21-point lead. Qualitative evaluation of the generated recipes confirms the advantages of our model.


\section{Related Work}

\noindent{\bf Large Foundational Models:}
The emergence of LLMs like BERT~\cite{kenton2019bert} and GPT-2~\cite{radford2019language} marked a significant leap in text understanding from summarization to reasoning. This success spurred exploration of even better LLMs and their application to visual-language tasks, including image captioning and visual question answering. Building on the success of LLMs like the 175B parameter model GPT-3.5~\cite{brown2020language}, recent smaller counterparts like Mistral~\cite{jiang2023mistral} and Phi-2~\cite{mojan2023phi2} demonstrate promising performance on various language tasks, suggesting potential benefits in efficiency and resource usage. Furthermore, recent proprietary models like GPT-4~\cite{achiam2023gpt} and BARD~\cite{bard_google_ai} have garnered significant attention for their multi-modal capabilities, but their proprietary nature restricts accessibility and computational feasibility. 

On the other hand, open-source multi-modal LLMs \cite{radford2021learning, li2023llava, awadalla2023openflamingo, li2023blip, zhu2023minigpt, dai2023instructblip} have demonstrated their effectiveness in various visual-language tasks. At the core of these multi-modal models lies a foundational LLM fine-tuned for understanding visual data. A common approach involves a pretrained vision-language encoder (e.g., CLIP~\cite{radford2021learning}) to extract visual features, which are then integrated with language embeddings through mapping layers \cite{liu2023visual, zhu2023minigpt} or cross-attention modules \cite{li2023blip, dai2023instructblip, awadalla2023openflamingo}. This approach has led to successful applications in domains like medicine \cite{moor2023med, li2023llava}, finance~\cite{wu2023bloomberggpt,liu2023fingpt}, and law~\cite{dahl2024large,anh2023impact}. While some research has explored applying these models to the food domain \cite{li2022food, yin2023foodlmm, chhikara2023fire, h2020recipegpt}, their performance remains limited due to ineffective or inadequate training strategies.

\medskip
\noindent{\bf Recipe Understanding:}
Early research in the food domain primarily focused on food image classification~\cite{he2021online, kaur2023deep}. Following this, interests shifted towards more intricate tasks including ingredient detection~\cite{chen2020study,rodriguez2023dining,chhikara2023fire}, recipe retrieval~\cite{chen2018deep, salvador2017learning, wahed2024fine}, ingredient substitution recommendations~\cite{pellegrini2021exploiting, li2022food}, and automatic recipe generation~\cite{chhikara2023fire, yin2023foodlmm, tanejamonte, farahani2023chef, bien2020recipenlg}. Notable attempts at recipe generation include Chef Watson's~\cite{varshney2019big} Bayesian network approach over a knowledge representation schema.  Wang et al.\ 
 \cite{wang2020structure} proposed a structure-aware generation method for recipes from food images. DoD~\cite{rodriguez2023dining} explored food recognition by learning fine-grained embeddings of food names and ingredients using BLIP-2~\cite{li2023blip} and Falcon 7B~\cite{almazrouei2022falcon}. RecipeGPT~\cite{h2020recipegpt} leveraged the GPT-2~\cite{radford2019language} architecture, while RecipeMC~\cite{tanejamonte} employed Monte Carlo Tree Search on top of GPT-2 for recipe generation. 

More recent works such as RecipeGM~\cite{reusch2021recipegm} and Chef Transformer~\cite{farahani2023chef} focused on generating recipes from pre-specified ingredient lists. FIRE~\cite{chhikara2023fire} utilizes BLIP ~\cite{li2022blip} model for food title generation and a ViT-based multi-class classifier for extracting ingredient lists, followed by the model T5~\cite{raffel2020exploring} for recipe generation. FoodLMM~\cite{yin2023foodlmm} fine-tuned LISA~\cite{lai2023lisa}, a multi-modal model, for diverse food-related tasks including classification, ingredient detection, segmentation and recipe generation.  While FoodLMM demonstrates improved performance across multiple tasks compared to baselines, its recipe generation capabilities remain a subject for further improvement.



\begin{figure}[!ht]
    \centering
    \includegraphics[scale=0.36]{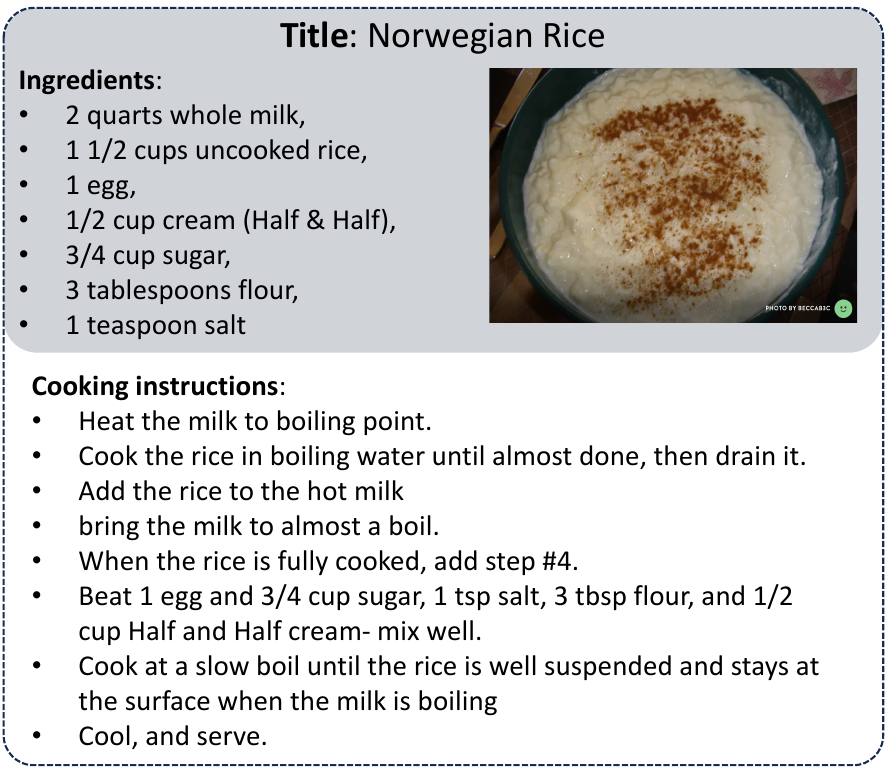}
    \caption{Sample recipe from the Recipe1M dataset. Title is denoted $X_t$, image $X_i$, ingredients $X_{ing}$, and instructions $X_{inst}$. 
    }
    \label{fig:sample_recipe}
    \Description[]{}
\end{figure}


\section{Visual Instruction-Following Data}
Building upon the success of LLaVA~\cite{liu2023visual} for visual instruction tuning, we adapt it to food recipe generation. Food recipes encompass both textual elements (title $X_t$, ingredients $X_{ing}$, and cooking instructions $X_{inst}$) and visual information (food image $X_i$), as illustrated in Figure \ref{fig:sample_recipe}. 
Despite several efforts to estimate cooking instructions from food images, none could produce good recipes compared to human.
Furthermore, a dearth of research exists regarding the generation of complete recipes solely from images, titles, ingredients, or combinations thereof. To bridge this gap, we develop instruction tuning prompts specifically designed to predict $Y_t$, $Y_{ing}$, $Y_{inst}$, or their combination. Our approach entails aligning food image embeddings with corresponding textual attributes by partially fine-tuning the model, followed by fine-tuning the complete model to estimate the desired food attributes through multi-modal fusion.

\begin{table*}[ht]
    \centering
    \begin{adjustbox}{width=0.9\textwidth}
    \begin{tabular}{|l|l|l|}
    \hline
      Input   & Output & sample prompt\\
      \hline
      \multicolumn{3}{|c|}{Stage 0 and 1 Training Prompts} \\
      \hline
       $X_i$  + $X_t$ + $X_{ing}$ & $Y_{inst}$ & Given <ingredients>, what are the key steps you need to follow to prepare a perfect <name>? \\
        $X_i$  + $X_t$ + $X_{ing}$ & $Y_{inst}$ & Please provide the step-by-step instructions for cooking a delicious <name> from scratch using the following ingredients: <ingredients>. \\
         $X_i$  + $X_t$ + $X_{ing}$ & $Y_{inst}$ &  Outline the steps to cook a <name> using ingredients: <ingredients> \\
    \hline
      \multicolumn{3}{|c|}{Stage 2 and 3 Training Prompts} \\
      \hline
       $X_i$  &  $Y_t$ &  What is the name of the dish in this image? \\
       $X_i$ + $X_{ing}$ & $Y_t$ & What is the name of the dish in this image? The ingredients used are: <ingredients> \\
       
       $X_i$   & $Y_{ing}$ &  Based on the features of the food in the image, provide a list of possible ingredients.  \\
       $X_i$   & $Y_{inst}$ & Describe how to prepare the meal shown in the image. \\
       $X_t$  & $Y_{inst}$ & Generate cooking instructions for <name>: \\
       $X_i$  + $X_t$  & $Y_{inst}$ & Generate cooking steps for <name> shown in this image. \\
       $X_i$  + $X_t$ + $X_{ing}$ & $Y_{inst}$ &  Elaborate on the steps involved in cooking <name> with these ingredients: <ingredients>\\
       $X_i$   & $Y_t$ + $Y_{ing}$ + $Y_{inst}$ &  Generate a name, ingredients, and cooking instructions for this dish: \\
       \hline
    \end{tabular}
    \end{adjustbox}
    \caption{Example prompts utilized at each training stage. We can see that S-0 and S-1 focus on generating cooking instructions, whereas S-2 and S-3 also on additional tasks. During training, we randomly select output task then we select input(s). 
    }
    \label{tab:exp2}
\end{table*}





\subsection{Food Concept Alignment Data}
To align food image embeddings with text embeddings, we randomly sample a question prompt $X_p$ for the generation of cooking instructions $Y_{inst}$ from the title $X_t$, ingredients $X_{ing}$, and the associated food image $X_i$. Sample prompts with placeholders are illustrated in Table~\ref{tab:exp2}.  The prompt $X_p$ contains placeholders tokens <$name$> and <$ingredients$> corresponding to the title $X_t$ and ingredients $X_{ing}$. During training, we substitute these placeholders with their actual values, resulting in the finalized prompt $X_q$. This refined prompt serves as the query for the model as demonstrated in Figure ~\ref{fig:model1}. Throughout the training, we structure inputs into a single-round instructions-following format, as exemplified below:

\begin{center}
    
    Human : $X_q$ $X_{i}$ <STOP> \textbackslash n 
    \\ Assistant : $Y_{inst}$ <STOP> \textbackslash n
    
\end{center}
  During training, optimization focuses solely on the layer that maps visual features to language embeddings. This targeted optimization aims to refine the visual embeddings and enhance their alignment with the food domain, ultimately improving the LLM's performance for recipe generation.

\subsection{Visual Instruction Tuning Data}
To adapt our model for food domain, we curated diverse prompts aimed at generating multiple textual attributes of a recipe from a food image and other textual attributes. These prompts effectively leverage the LLM's ability to perform multi-modal text generation. Specifically, each prompt was designed to elicit a targeted output from the LLM. For instance, one prompt instructed the model to generate the food name based solely on its image. Another prompt tasked the model with predicting the cooking instructions, utilizing both the food image and the provided name. We employed GPT-3.5 to generate prompts for the following target outputs: food name ($Y_t$), cooking instructions ($Y_{inst}$), and cooking ingredients ($Y_{ing}$). Examples of these prompts are presented in Table~\ref{tab:exp2}. During training, we randomly select a task and then a prompt specific to the selected task. The chosen prompt either may demand the prediction of a single output (title, ingredients, or instructions) or multiple outputs from the provided inputs. In cases where the recipe lacks an associated image, an empty image is utilized.

Our multi-stage fine-tuning process progressively enhances the model's understanding of food recipes. Initially (Stage-0), visual embeddings are projected into the language domain, establishing a foundation for subsequent learning. Stage-1 focuses on recipe comprehension by training the model to generate cooking instructions based on the provided food image, title, and ingredients. Subsequent stages (Stages-2 and Stage-3) increase task complexity and reduce input information to promote deeper recipe knowledge acquisition. In the cooking instruction task, diverse prompts expose the model to varying input modalities (image-only, title-only, image-title, and image-title-ingredients), fostering robustness in recipe generation. Finally, the model is also challenged to predict recipe title, ingredients, and cooking instructions solely from the image, solidifying its ability to infer comprehensive recipe information from limited visual input.

\section{LLaVA-Chef: Adapting LLaVA to food domain}

The performance of LLaVA-Chef is gradually improved by a meticulously designed multi-stage training strategy to unlock its full potential as described below in detail.

\subsection{Stage 0: Food domain adaptation}
To bridge the gap between visual and language modalities, LLaVA leverages a linear layer to project visual features into the language space. In Stage-0, we concentrate on fine-tuning the mapping layer using image-recipe pairs from the Recipe1M dataset \cite{salvador2017learning}. As illustrated in Figure \ref{fig:model1}, the food image $X_i$, name $X_t$ and ingredients $X_{ing}$ are input to the model and the model is asked to generate cooking instructions. Optimization of the mapping layer is achieved through the standard cross-entropy loss function defined as follows: 
\begin{equation}
    L_{CE} = CE (p (Y_{inst}),  p(\hat{Y}_{inst})) 
    \label{eq:loss_s0}
\end{equation}
Where, $p(Y_{inst})$ is probability of ground truth cooking instruction as one hot-vector, $p(\hat{Y}_{inst})$ indicates probability of the cooking instructions predicted by the model. This fine-tuning aims to optimize the alignment of the visual embeddings with their corresponding language representations, enhancing the model's ability to capture the nuances of visual information relevant to recipes. Note that this step fine-tunes the mapping layer to better understand the food images.

\subsection{Stage 1: Learning the language of recipes}
To train our model on predicting cooking instructions from image, title, and ingredients, we curated a dataset of 35 unique prompts. Each prompt incorporates special tokens: <name> representing the food title and <ingredients> signifying the listed ingredients. During training, we randomly sample a prompt, then replace these special tokens with the title and ingredients of the recipe and fine-tune the entire backbone LLM model. This approach allows the model to learn food-domain embeddings from both visual and textual data seamlessly. Recognizing that not all recipes may have accompanying images, we employed a strategy for handling missing visuals. When an image is unavailable, we substitute it with a black (empty) image as a placeholder. This enables the model to learn from the remaining textual attributes (title and ingredients) and still estimate cooking instructions even without image input. The model is optimized using the default cross entropy loss function as defined above in equation ~\ref{eq:loss_s0}.

\subsection{Stage 2: Boosting model adaptability via prompt diversity}
The Recipe1M dataset~\cite{salvador2017learning} offers four attributes for each recipe: image, title, ingredients, and cooking instructions. While image contributes visual information, the latter three act as textual attributes. To diversify our training prompts, we expanded our initial set of 35 prompts by utilizing GPT-3.5 to generate prompts for various recipe-related tasks, bringing the total to 102 prompts, some examples are shown in Table~\ref{tab:exp2}. These prompts are task-specific, explicitly defining the input and target output for each prediction scenario. During training, we randomly select a task (what to predict) and a corresponding prompt. We opted to retain cross-entropy as our chosen loss function. This approach fosters model generalizability, enabling it to predict the desired output (e.g., title, ingredients, or instructions) from image, title, or ingredients via fine-tuning as shown in Figure ~\ref{fig:model1}. To further improve generalization, we adopted a strategy where at most $50\%$ of the ingredients are omitted from the input during training. This forces the model to infer missing ingredients based on the remaining information, ultimately leading to improved performance across all tasks, including cooking instruction generation from solely image or title.

\subsection{Stage 3: Optimizing the recipe language}
To enhance the language quality and achieve predictions closer to the ground truth, we extended the training of our model from Stage-2 by introducing an additional penalty loss, based on the commonly used BLEU~\cite{papineni2002bleu} and Rouge~\cite{lin2004rouge} scores that were initially formulated to evaluate machine translation and text summarization tasks. However, one cannot directly optimize these metrics as additional loss terms, since they are non-differentiable (e.g., they are based on $n$-gram counts). Instead of optimizing them directly, we propose a novel formulation where we use the scores as a multiplicative or scaling factor for the cross-entropy loss. Let $Y_{label}$ denote the ground truth recipe, $Y_{pred}$ the generated recipe (note: $label$ can refer to any of the inputs such as title, image, ingredients and/or cooking instructions). Next, define $L_{bleu} = 1 - BLEU(Y_{label}, Y_{pred})$ as the penalty from the SacreBLEU score~\cite{post2018call}, and $L_{rougeL} = 1 - rougeL(Y_{label}, Y_{pred})$ as the penalty from Rouge-L~\cite{lin-och-2004-automatic}. Since higher scores are better (with 1 being the maximum score), we penalize by subtracting them from 1. We then combine both into a joint scaling penalty:
\begin{equation}
    L_{BR} = \lambda_{bleu} (1- L_{bleu}) +  \lambda_{rougeL} (1- L_{rougeL})
    \label{eq:alpha}
\end{equation}
where $\lambda_{bleu}$ and $\lambda_{rougeL}$ are weighting factors.
Next, we multiply the (per-sample) scaling penalty $L_{BR}$ with the cross-entropy loss $(L_{CE})$, as follows:
\begin{equation}
    L_{final} = L_{BR} \times L_{CE} 
    \label{eq:loss}
\end{equation}
As such $L_{BR}$, while non-differentiable, works as a (per sample) scaling constant, thus scaling and penalizing the overall loss when the value of either of the metrics goes down; the final loss remains differentiable.
 This multi-objective approach holds the promise of generating more fluent, accurate, and semantically aligned recipe instructions, as we investigate in the following section. 



\begin{table*}[!ht]
    \centering
 \begin{adjustbox}{width=0.98\textwidth}
    \begin{tabular}{|l|l|c|c|c|c|c|c|c|c|c|c|c|}
\hline

Method & Inputs & BLEU-1 & BLEU-2 & SacreBLEU & METEOR & ROUGE-1 & ROUGE-2 & ROUGE-L  & CIDEr & Perplexity $\downarrow$ \\
\hline
Chef Transformer~\cite{farahani2023chef} & $X_{ing}$ & 0.271 & 0.128 & 0.037 & 0.117 &0.259 & 0.057 & 0.133 & 0.046 & 54.21   \\
\hline
GPT-2~\cite{radford2019language}& $X_{t}$ + $X_{ing}$ & 0.084 & 0.032 & 0.01 & 0.037 & 0.111 & 0.018 & 0.088 & 0.01 & 2.15   \\
Mistral~\cite{jiang2023mistral} & $X_{t}$ + $X_{ing}$ & 0.126 & 0.072 & 0.04 & 0.079 & 0.179 & 0.055 & 0.106 & \textbf{0.05} & 26.45    \\
Phi-2~\cite{mojan2023phi2} & $X_{t}$ + $X_{ing}$ & 0.143 & 0.07 &  0.027 & 0.147 & 0.202 & 0.047 & 0.108 & 0.002 &  32.41   \\
LLaMA~\cite{touvron2023llama} & $X_{t}$ + $X_{ing}$ & 0.234 & 0.119 & 0.049  & 0.16 & 0.29 & 0.075 & 0.155 & 0.043 & 2.86 \\
\hline
InstructBLIP-T5xl~\cite{dai2023instructblip} & $X_{i}$ + $X_{t}$ + $X_{ing}$ & 0.014 & 0.006 & 0.001  & 0.037 & 0.137 & 0.022 & 0.094 & 0.014 & 68.71  \\
InstructBLIP-Vicuna~\cite{dai2023instructblip} & $X_{i}$ + $X_{t}$ + $X_{ing}$ & 0.0013 & 0.0004 & 0.0001  & 0.026 & 0.103 & 0.012 & 0.069 & 0.004 & 135.75  \\

MiniGPTv2~\cite{chen2023minigpt} & $X_{i}$ + $X_{t}$ + $X_{ing}$ & 0.232 & 0.115 & 0.06  & 0.139 & 0.257 & 0.06 & 0.135 & 0.03 & 157.0  \\

MiniGPT4-LLaMA-2~\cite{zhu2023minigpt} & $X_{i}$ + $X_{t}$ + $X_{ing}$ & 0.2754 & 0.141 & \textbf{0.07}  & \textbf{0.204} & 0.353 & 0.094 & 0.173 & 0.032 & 11.78  \\

LLaVA~\cite{liu2023visual} & $X_{i}$ + $X_{t}$ + $X_{ing}$ & \textbf{0.29} & \textbf{0.155} & 0.06 & 0.2 & \textbf{0.366} & \textbf{0.105} & \textbf{0.184} & 0.041 & \textbf{2.6} \\

\hline

\end{tabular}
\end{adjustbox}
\caption{
Performance of pretrained foundational models on our $test1k$. Notably, pretrained LLaVA, outperforms other evaluated models on most metrics, showcasing its ability to generate food recipes.
}
\label{tab:test1k_llms}

    \centering
 \begin{adjustbox}{width=0.9\textwidth}
    \begin{tabular}{|l|c|c|c|c|c|c|c|c|c|c|c|c|}
\hline

Method & Inputs & BLEU-1 & BLEU-2 & BLEU-3 & BLEU-4 & SacreBLEU & METEOR & ROUGE-1 & ROUGE-2 & ROUGE-L  & CIDEr & Perplexity $\downarrow$ \\
\hline
Chef Transformer~\cite{farahani2023chef} &  $X_{ing}$ & 0.267 & 0.127 &0.064 &0.034& 0.038 & 0.116 & 0.262 & 0.059 & 0.136 & 0.045 & 30.62 \\
Mistral~\cite{jiang2023mistral} & $X_t$ + $X_{ing}$ & 0.130 & 0.075 & 0.048 & 0.033 & 0.041 & 0.082 & 0.188 & 0.058 & 0.111  & 0.063 & 75.36    \\
LLaMA~\cite{touvron2023llama} & $X_t$ + $X_{ing}$ & 0.252 & 0.129 &0.072 & 0.043 & 0.053 & 0.156 & 0.293 & 0.077 & 0.156 & 0.031 & 2.86   \\

LLaVA~\cite{liu2023visual} & $X_i$ + $X_t$ + $X_{ing}$ & 0.297 & 0.159 &0.089 & 0.042 & 0.061 &  \textbf{0.2} & 0.368 & 0.106 & 0.183 & 0.037 &  2.92   \\
\hline
LLaVA-Chef-S1 & $X_i$ + $X_t$ + $X_{ing}$ & 0.322 & 0.19 &0.117 &0.075 & 0.096 & 0.159 & 0.404 & 0.141 & 0.217 &  \textbf{0.187} & 2.62  \\
LLaVA-Chef-S2 & $X_i$ + $X_t$ + $X_{ing}$ & 0.331 & 0.193 & 0.118 &0.075 & 0.09 & 0.159 & 0.396 & 0.136 & 0.213 &  0.176 & 2.86  \\
LLaVA-Chef-S3 & $X_i$ + $X_t$ + $X_{ing}$ & \textbf{0.362} & \textbf{0.215} & \textbf{0.135} & \textbf{0.089} & \textbf{0.167} & 0.188 & \textbf{0.473} & \textbf{0.172} & \textbf{0.241} &  0.216 &  \textbf{2.38} \\
\hline
\end{tabular}
\end{adjustbox}
\caption{Results on Recipe1M test set $test50$K (randomly selected 50,507 test samples, fixed for all models). Our model, LLaVA-Chef, gradually improves from Stage-1 to Stage-3 on almost all the metrics.}

\label{tab:test50k}
\end{table*}


\section{Experiments}

\subsection{Experimental setup}

\paragraph{\textbf{Dataset:}}  We leveraged Recipe1M~\cite{salvador2017learning}, a large-scale recipe dataset boasting $1$ million recipes and 819,000 food images. Each recipe comprises a title, ingredients list, and cooking instructions, with several samples also accompanying one or more images. Recipe1M already provides training, validation, and test splits. For the training phase, we utilized the entire training set consisting of 720,639 recipes (with 619,508 images). However, during testing, we focused on recipes with at least one image. After cleaning the test set by removing samples lacking images or containing corrupted ones, we obtained two curated testing subsets:
\begin{itemize}
    \item \textbf{ $test50k$: } All 50,507 test samples from Recipe1M test that contain at least one image. 
    \item $test1k$: We selected another 1,000 samples (randomly) as $test1k$ set for detailed qualitative analysis.
\end{itemize}
\paragraph{\textbf{Metrics:}}
To evaluate the generated text quality compared to the ground truth, we employed several image caption and language translation metrics. These metrics include BLEU~\cite{papineni2002bleu}, a precision-based metric specifically designed for machine translation, Rouge~\cite{lin2004rouge}, a recall-oriented metric for text summarization, METEOR~\cite{elliott2013image} and CIDEr~\cite{vedantam2015cider}, which were specifically developed for assessing image caption quality and exhibit strong correlation with human subjective judgments. Perplexity~\cite{jelinek1977perplexity}, a measure of language model uncertainty, was also included to provide additional insights into fluency and coherence of the generated text.

\paragraph{\textbf{Model Training:}} 
Our model, LLaVa-Chef, was trained in four consecutive stages on four NVIDIA RTX A6000 48G GPUs with a batch size of 32. We set learning rate to 2e-5 with a cosine learning scheduler at a warmup ratio of 0.03. Stages 0, 1 and 2 employed the standard cross-entropy loss function. In Stage-3, loss was scaled based on BLEU ($\lambda_{bleu} = 1.01$), and Rouge-L ($\lambda_{rougeL} = 1$). This multi-objective approach prioritized language quality, ultimately leading to improved performance in generated text fidelity when compared to ground-truth recipes. Our model and data is publicly available at \url{https://github.com/mohbattharani/LLaVA-Chef}.

\subsection{LLaVA fine-tuning }
Our investigation into recipe generation compared multiple high-performing open-source general-purpose LLMs. We also evaluated Chef Transformer~\cite{farahani2023chef} (T5~\cite{raffel2020exploring} fine-tuned on the RecipeNLG dataset~\cite{bien2020recipenlg}), the sole publicly available open-source recipe generation model at the time (December 2023). Evaluation on a 1000 sample test set ($test1k$) drawn from the Recipe1M dataset (as detailed in Table~\ref{tab:test1k_llms}) revealed LLaVA~\cite{liu2023visual}, a multi-modal LLM, to outperform all contenders, including Chef Transformer. Consequently, LLaVA was chosen for further analysis and fine-tuned on the Recipe1M dataset for enhanced performance. Our training protocol employed a multi-stage fine-tuning approach. Initially, during Stage-0, we conducted fine-tuning for the projection layer over the course of two epochs. Subsequently, throughout the remaining three stages (Stage 1-3), the entire model was fine-tuning for two epochs in each stage.

Our analysis of current open-source LLMs presented in Table~\ref{tab:test1k_llms} reveals intriguing performance in the food domain. In case of text-only models, Chef-Transformer shown higher BLEU-1 and BLEU-2 scores but it has lower scores on SacreBLEU, METEOR, and Rouge-L than LLaMA, indicating potential trade-offs in generation quality. Whereas, comparing all the models, LLaVA seems to outperform. The higher perplexity scores suggests that, with the exception of LLaMA, MiniGPT-4 and LLaVA, all models struggle to generate good quality language, potentially generating text exhibiting hallucinations or incomplete sentences. Though Mistral has impressive performance on standard benchmarks, its higher perplexity score and scores for other metrics lower than Phi-2 raises questions about its effectiveness in this specific context. InstructBLIP \cite{dai2023instructblip} generated recipes for more like caption rather than cooking steps. The training data of MiniGPT-4~\cite{zhu2023minigpt} contains food images paired with cooking instructions, hence it is comparable to Chef-Transformer for recipe generation on several metrics. Overall, LLaVA stands out, achieving remarkable performance on most metrics. 


\subsection{Quantitative Results}
The results presented in Table~\ref{tab:test1k_llms} for $test1k$ set demonstrate that the pre-trained LLaVA~\cite{liu2023visual} outperforms other LLMs including Chef-Transformer~\cite{farahani2023chef}, despite Chef-Transformer being trained on recipe dataset. A similar trend was also found on test set $test50k$ as shown in Table~\ref{tab:test50k}, comparing top 4 row, LLaVA has higher scores. 
Our LLaVA-Chef model therefore extends the baseline LLaVA model via our novel multi-stage training and fine-tuning framework outlined above. Notably, our model, LLaVA-Chef outperforms other models, with its BLEU and Rouge scores indicating the alignment of the generated cooking instructions with the ground truth.

\paragraph{\textbf{Open source LLMs:}}

Due to limited benchmarks for recipe generation, we explored the performance of prominent LLMs on the Hugging Face Leader board (December 20, 2023). These include well-established models like GPT-2~\cite{radford2019language}, and LLaMA~\cite{touvron2023llama}, as well as recent high-performing options such as Mistral (7B parameters)~\cite{jiang2023mistral} and Phi-2 ~\cite{mojan2023phi2}. 
We also considered four multi-modal models in our study including InstructBLIP~\cite{dai2023instructblip}, MiniGPTv2~\cite{chen2023minigpt}, MiniGPT-4~\cite{zhu2023minigpt} and LLaVA~\cite{liu2023visual} due to their exceptional performance on visual-language tasks. Additionally, we evaluated Chef Transformer~\cite{farahani2023chef}, a fine-tuned T5~\cite{raffel2020exploring} model specifically designed for recipe generation, offering an open-source option for comparison.


\paragraph{\textbf{Comparison with existing methods:}}

Direct comparison with the existing literature is challenging due to discrepancies in reported results and limited dataset accessibility. The partial availability of Recipe1M dataset and outdated URLs hinder consistent evaluation. For examples RecipeMC~\cite{tanejamonte} is evaluated on $1000$ samples from Recipe1M dataset but they did not share those samples. Similarly, FIRE~\cite{chhikara2023fire} could get $56K$ samples from test set of Recipe1M dataset as few URL were no more accessible. In our case, we could get only $50,507$ test samples that contain at least one image per recipe. Although, the test set used by baseline methods and ours might be slightly different, the scores give us a general idea about the performance of the models.


\begin{table}[!ht]
    \centering
     \begin{adjustbox}{width=0.37\textwidth}
    \begin{tabular}{|l|c|c|}
    \hline
        Method & SacreBLEU & ROUGE-L \\
        \hline
       Inverse Cooking\cite{salvador2019inverse} & 0.055 & 0.195 \\
       FIRE~\cite{chhikara2023fire} & 0.06 & 0.212 \\
       FoodLMM~\cite{yin2023foodlmm} & 0.062 & \textbf{0.369} \\
       Chef Transformer \cite{farahani2023chef}  & 0.046 & 0.175 \\
       Chef Transformer$^*$ & 0.038 & 0.136 \\
       LLaVA$^*$  & 0.061 & 0.183 \\
       \hline
       LLaVA-Chef-S3$^*$ (Our) & \textbf{0.158} & 0.228\\
       \hline
    \end{tabular}
    \end{adjustbox}
    \caption{
    Results on Recipe1M test set: Due to inconsistency in datasets and lack of publicly available models, results based on our $test50k$ benchmark dataset are marked with $*$. 
    }
    \label{tab:sota1}
    \centering
 \begin{adjustbox}{width=0.45\textwidth}
    \begin{tabular}{|l|c|c|c|c|c|}
\hline

Method  &  Perplexity (gt/pred) $\downarrow$ &  ROUGE-1 & ROUGE-2 & BLEU   \\
\hline
RecipeMC~\cite{tanejamonte}  &   2.934 / 7.337 & 0.362 & 0.115 & 0.08 \\
LLaVA~\cite{liu2023visual} &   6.8 / 2.6 & 0.367 & 0.105  & 0.06 \\
\hline
LLaVA-Chef-S3 & 4.14 / 2.4 & \textbf{0.473} & \textbf{0.17}  & \textbf{0.17} \\

\hline
\end{tabular}
\end{adjustbox}
\caption{Results on 1000 test recipes from Recipe1M dataset (gt: ground truth, pred: predicted or generated text). RecipeMC test recipes are taken from \cite{tanejamonte}.
}
\label{tab:sota2}
\end{table}


Our model in general outperforms the baseline methods as evident in Table~\ref{tab:sota1} and Table ~\ref{tab:sota2}. We took the scores for Chef Transformer~\cite{farahani2023chef}, Inverse Transformer~\cite{salvador2019inverse}, FIRE~\cite{chhikara2023fire}, and FoodLMM~\cite{yin2023foodlmm} from their respective publications. Additionally, we conducted an evaluation of the publicly available Chef Transformer~\cite{farahani2023chef} on our $test50k$ set. Intriguingly, our evaluation yielded lower scores for Chef Transformer compared to those reported in its original publication. Notably, the pretrained general-purpose LLaVA~\cite{liu2023visual} marginally surpassed FIRE and is close to FoodLMM in terms of SacreBLEU score. Despite being built upon LLaVA, FoodLMM~\cite{yin2023foodlmm} only achieved a 1-point improvement in SacreBLEU score, although its Rouge-L score is significantly higher.

On the other hand, our model, LLaVA-Chef, as seen in Table~\ref{tab:sota1} demonstrates superior performance, achieving a remarkable nearly 10-point margin over other models in SacreBLEU score, even with second best Rouge-L score. As shown in Table~\ref{tab:sota2}, LLaVA-Chef surpasses RecipeMC on both Rouge and BLEU scores. This significant performance gain validates the effectiveness of our approach.


\begin{table*}[ht]
    \centering
 \begin{adjustbox}{width=0.9\textwidth}
    \begin{tabular}{|l|l|c|c|c|c|c|c|c|c|c|c|c|}
\hline

Cuisine & Train Samples & BLEU-1 & BLEU-2 & BLEU-3 & BLEU-4 & SacreBLEU & METEOR & ROUGE-1 & ROUGE-2 & ROUGE-L  & CIDEr & Perplexity $\downarrow$ \\
\hline
North-American  & 13119 & 0.377  & 0.216  & 0.13  & 0.082  & 0.127  & 0.185   & 0.445  & 0.133  & 0.198  & 0.202 & 16.36 \\
American   & 9026 & 0.376  & 0.215  & 0.129  & 0.081  & 0.125  & 0.184    & 0.411  & 0.124  & 0.183 & 0.2 & 18.79 \\
European & 5683 & 0.384 &  0.22 &  0.133 &  0.084 &  0.143 &  0.187  &  0.411 &  0.125 &  0.204 &  0.198 &  18.60 \\
Asian &2526 & 0.377 &  0.215 &  0.129 &  0.08 &  0.152 &  0.184 &   0.511 &  0.172 & 0.232 &  0.194 & 10.46 \\
Mexican & 2472 & 0.379 &  0.218 &  0.131 &  0.083 &  0.131 &  0.186 &    0.507 &  0.15 &  0.224 &  0.205 &11.9 \\
Italian&2047 & 0.393 &  0.226 &  0.137 &  0.087 &  0.214 &  0.19 &    0.526 &  0.194 &  0.273 &  0.193 &9.57 \\
Indian & 544& 0.377 & 0.215 &  0.128 &  0.081 &  0.104 &  0.184 &   0.531 &  0.173 &  0.229 &  0.193 & 11.34  \\
French &  427& 0.377 &  0.215 &  0.129 &  0.081 &  0.04 &  0.184 &   0.097 &  0.011 &  0.043 &  0.191 & 15.29 \\
English & 248 & 0.377 &  0.216 &  0.13 &  0.083 &  0.106 &  0.185 &   0.241 &  0.038 &  0.148 &  0.226 & 16.42 \\
Middle-Eastern & 267 & 0.376 &  0.215 &  0.128 &  0.08 &  0.134 &  0.184 &    0.487 &  0.103 &  0.218 & 0.197 & 11.43\\
Thai & 252 & 0.374 &  0.214 &  0.127 &  0.08 &  0.315 &  0.183 &   0.664 &  0.311 &  0.28 &  0.188 & 8.56 \\
German &247&  0.378 &  0.216 &  0.129 &  0.081 &  0.068 &  0.185 & 0.366 &  0.091 &  0.173 &   0.192 &  42.98\\
Russian & 223&  0.377 &  0.215 &  0.129 &  0.081 &  0.135 &  0.184 &   0.459 &  0.106 &  0.182 &  0.189 & 8.71\\
Japanese &132 &  0.38 &  0.218 &  0.13 &  0.082 &  0.055 &  0.186 &   0.362 &  0.101 &  0.2 &  0.198 & 10.48\\

\hline
test1k &-& 0.366  & 0.218  & 0.137  & 0.09  & 0.17  & 0.189  & 0.473  & 0.17  & 0.24  & 0.242  & 17.9 \\
\hline

\end{tabular}
\end{adjustbox}
\caption{Performance of LLaVA-Chef on generating recipe that belong to different cuisines}
\label{tab:cusines}
\end{table*}

\paragraph{\textbf{Performance on different cuisines:}}
To evaluate the generalization of LLaVA-Chef, we report the performance of our model on test samples from different cuisines in Table \ref{tab:cusines}, and compare with scores on test1k. For most of the cuisines, BLEU and Rouge scores are almost same. Our model shows lowest Rouge scores for French and higher perplexity for German. In general, most of the scores are close to the overall scores on $test1k$ set indicating the the model generalizes across cuisines, even for those with few training examples (e.g., Japanese or Russian).



\subsection{Qualitative Results}
Beyond quantitative metrics, evaluating the qualitative aspect of generated recipes is crucial. Figure~\ref{fig:result_samples1} presents two recipes generated by Chef-Transformer, LLaVA and LLaVA-Chef (Ours). In the left-hand example, all models recommend a lower \textit{temperature} than the ground truth, but the \textit{baking time} remains consistent. In the right-hand example, all models suggest the same \textit{oven temperature} but vary in recommended \textit{cooking time}. LLaVA-Chef generates concise recipes with high accuracy, often surpassing other models and even the ground truth in clarity. When manually looking at the generated recipe, we observe that GPT-2, Mistral and Phi-2 struggle to produce a cohesive recipe, Chef Transformer generated recipe do not have sufficient information, LLaMA sometime fails to generate correct recipes, and InstructBLIP generates text which looks like a caption rather than cooking steps. LLaVA generates detailed recipes but hallucination is common in generated text. However, our LLaVA-Chef generated recipe is concise and closely resembles human generated ground truth recipe. 

\begin{figure*}[ht]
\begin{center}
\begin{adjustbox}{width=0.88\textwidth}
  \centering
  \fontsize{7.pt}{7.pt}\selectfont
    \begin{tabularx}{\textwidth}{lX}
    \toprule
    
         \raisebox{-\totalheight}{\includegraphics[width=0.08\textwidth, height=1.2cm]{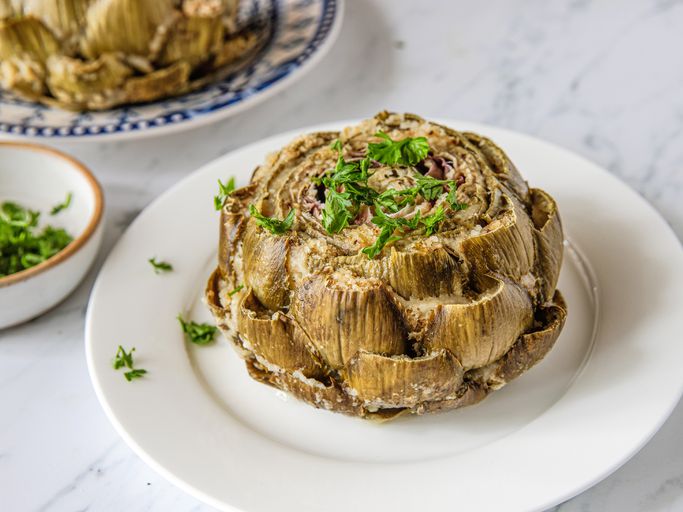}}
         &
    \textbf{Prompt:} Generate a comprehensive recipe for crafting \textbf{Stuffed Artichokes} using the following ingredients: \textit{2 lbs lean ground beef, 7 ounces fresh mushrooms chopped, 1 large onion, diced finely, 2 large eggs, 12 cup celery, diced finely, 12 cup tomato puree, 12 cup water, 1 teaspoon salt to taste, 12 teaspoon freshly ground black pepper, 1 teaspoon thyme, 3 cloves garlic, minced, 24 large canned artichoke bottoms or 24 large frozen artichoke bottoms, 2 cups beef broth}.
    \\
    \midrule 
     Ground Truth &
      1. Preheat the oven to 350 degrees F.$\backslash$n2. Combine all the ingredients except for the artichoke bottoms and the beef broth.$\backslash$n3. Divide and mound the meat mixture on the artichoke bottoms, about 4 tablespoons each.$\backslash$n4. In a well greased baking pan, place the filled artichokes, filled side up, in one layer.$\backslash$n5. Pour the broth over the whole thing.$\backslash$n6. Bake, lightly covered with foil, for 1 1/2 hours.$\backslash$n7. Uncover and bake another 10-15 minutes, until browned.$\backslash$n8. Serve hot.
    \\ 
     \midrule
    \textbf{LLaVA-Chef-S3} &
    1. Preheat oven to 350 degrees F.$\backslash$n2. In a large skillet, saute the ground beef, mushrooms, onion, and garlic until the meat is browned.$\backslash$n3. Drain off the fat.$\backslash$n4. In a large bowl, combine the meat mixture with the eggs, celery, tomato puree, water, salt, pepper, and thyme.$\backslash$n5. Stuff the artichoke bottoms with the meat mixture.$\backslash$n6. Place the stuffed artichokes in a large baking dish.$\backslash$n7. Pour the beef broth over the artichokes.$\backslash$n8. Cover the dish with aluminum foil and bake for 1 hour.$\backslash$n9. Remove the foil and bake for an additional 15 minutes.
     \\
     \midrule
    LLaVA-Chef-S2 & 1. Preheat oven to 350 degrees F.$\backslash$n2. In a large bowl, combine the ground beef, mushrooms, onion, celery, eggs, tomato puree, water, salt, pepper, and thyme.$\backslash$n3. Mix well.$\backslash$n4. Stuff the artichoke bottoms with the mixture.$\backslash$n5. Place the artichoke bottoms in a large baking dish.$\backslash$n6. Pour the beef broth over the artichokes.$\backslash$n7. Cover the dish with aluminum foil and bake for 45 minutes.$\backslash$n8. Remove the foil and bake for an additional 15 minutes.
    \\
    \midrule
    LLaVA-Chef-S1 & 1. Preheat oven to 350 degrees F.$\backslash$n2. In a large skillet, brown the ground beef with the onion and garlic.$\backslash$n3. Add the mushrooms and cook until softened.$\backslash$n4. Drain the artichoke bottoms and add them to the skillet.$\backslash$n4. Stir in the tomato puree, water, salt, pepper and thyme.$\backslash$n5. Bring to a boil, reduce heat, and simmer for 15 minutes.$\backslash$n6. Place the artichoke bottoms in a baking dish and cover with the meat mixture.$\backslash$n7. Bake for 30 minutes.$\backslash$n8. Serve with rice or pasta.
     \\
     \midrule 
     LLaVA &
     To prepare a delicious Stuffed Artichokes, follow these steps:$\backslash$n$\backslash$n1. Preheat your oven to 375°F (190°C).$\backslash$n2. In a large bowl, combine the lean ground beef, mushrooms, onion, and eggs. Season the mixture with salt, pepper, and thyme to taste.$\backslash$n3. Add the celery, tomato puree, water, and minced garlic to the bowl. Mix everything together until well combined.$\backslash$n4. Place the mixture in a large baking dish and bake for about 45 minutes, or until the filling is cooked through and the edges are slightly browned.$\backslash$n5. While the filling is cooking, prepare the artichoke bottoms. If using fresh artichoke bottoms, trim the ends and remove the tough outer leaves. If using frozen artichoke bottoms, follow the package instructions for preparation.$\backslash$n6. After the filling is cooked, carefully remove it from the oven and let it cool for a few minutes.$\backslash$n7. Carefully stuff the cooked filling into the artichoke bottoms, dividing the mixture evenly among the artichokes.$\backslash$n8. Pour the beef broth over the stuffed artichokes, making sure to cover them completely.$\backslash$n9. Return the baking dish to the oven and bake for another 25-30 minutes, or until the artichokes are tender and the filling is hot and bubbly.$\backslash$n10. Serve the Stuffed Artichokes hot, garnished with fresh parsley if desired.$\backslash$n$\backslash$nEnjoy your delicious Stuffed Artichokes!
     \\
   
    \bottomrule
    \end{tabularx}
\end{adjustbox}
  
\end{center}
\caption{Example recipes generated by pre-trained LLaVA and each stage of our model. We can see how each stage successively improves the generated recipe, showcasing the effectiveness of our multi-stage training.   
}
  \label{fig:pred_stages1}
  \Description[]{}
\end{figure*}

We also look at how our LLaVA-Chef's multi-stage approach successively improves the generated recipes. We found that Stage-1 exhibits minor discrepancies, while Stage-3 generates accurate recipes with correct ingredients (see Figure~\ref{fig:pred_stages1}). Further analysis reveals that sometimes the recipes are semantically equivalent but linguistically different causing lower scores compared to the ground-truth. Finally, we looked at the impact of combinations of food image $X_i$, title $X_t$, ingredients $X_{ing}$ as inputs to our model. We find that solely relying on the image sometimes makes dish prediction difficult, leading to a flawed recipe, though high quality images can provide good results. Providing the title significantly improves the generation. While LLaVA-Chef achieves promising results on the Recipe1M dataset, certain limitations emerged upon closer examination. To summarize, some recipes 
closely resemble the corresponding ground-truth recipes, while others exhibit significant linguistic divergence resulting in lower Rouge-L scores even though generated recipes are semantically equivalence with the ground truth. For instance, a single step of the ground truth recipe is sometimes split into several steps in the generated recipes, conveying the same information but with different phrasing.

\begin{figure*}[htbp]
\begin{center}
\begin{adjustbox}{width=0.9\textwidth}
  \centering
  \fontsize{7.0pt}{7.0pt}\selectfont
    \begin{tabularx}{\textwidth}{XX}
    \toprule
    \multicolumn{1}{c}{\textbf{Ground Truth}}  & \multicolumn{1}{c}{\textbf{LLaVA-Chef}} \\
    \toprule
    \multicolumn{2}{>{\hsize=\dimexpr 2\hsize+\arrayrulewidth}X}{
    \cellcolor{gray!30} \textbf{1. Prompt:} Compose a recipe for \textbf{Avocado-Hummus Dip} incorporating the following ingredients: \textit{2 medium Hass avocados, peeled and chopped, One 15-ounce can chickpeas, drained and rinsed, 1/4 cup fresh lemon juice, 1 1/2 tablespoons tahini (sesame paste), 1/2 cup extra-virgin olive oil, plus more for drizzling, Kosher salt, Freshly ground pepper, Assorted crudites, bread and tortilla chips, for serving.}
    }  \\ 
    \midrule
    1. In a food processor, puree the avocados with the chickpeas, lemon juice and tahini.$\backslash$n2. Add the 1/2 cup of oil and puree until smooth; season with salt and pepper.$\backslash$n3. Transfer the dip to a bowl, drizzle with olive oil and serve with crudites, bread and chips. &
    1. In a food processor, combine the avocados, chickpeas, lemon juice, tahini, olive oil, and salt and pepper.$\backslash$n2. Process until smooth and creamy.$\backslash$n3. Transfer to a serving bowl and drizzle with more olive oil.$\backslash$n4. Serve with crudites, bread slices, and tortilla chips. \\
    \midrule
    
    \multicolumn{2}{>{\hsize=\dimexpr 2\hsize+\arrayrulewidth}X}{
    \cellcolor{gray!30} \textbf{2. Prompt:} Generate a comprehensive recipe for crafting \textbf{Anchovy and Caper Dressing} with the listed ingredients: \textit{2 tablespoons lemon juice, 5 tablespoons olive oil, 1 teaspoon fresh ground black pepper, 6 anchovy fillets, pounded to a coarse paste, 1 tablespoon capers, chopped.}
    }  \\ 
    \midrule 

    1. Mix together all ingredients.$\backslash$n2. Salt should not be needed because of the anchovies. &
    1. In a small bowl, whisk together the lemon juice, olive oil, and pepper.$\backslash$n2. Add the anchovy paste and capers.$\backslash$n3. Whisk again to combine.$\backslash$n4. Taste and adjust seasonings if necessary.$\backslash$n5. The dressing can be refrigerated for up to 2 days. \\
    \midrule
    
    \multicolumn{2}{>{\hsize=\dimexpr 2\hsize+\arrayrulewidth}X}{
    \cellcolor{gray!30} \textbf{3. Prompt:} Given \textit{1 1/2 cups buttermilk, 1 1/2 tablespoons margarine, 2 tablespoons white sugar, 1 teaspoon salt, 3 cups bread flour, 1 1/3 cups whole wheat flour, 2 1/4 teaspoons active dry yeast}, what are the key steps you need to follow to prepare a perfect \textbf{Buttermilk Bread} ?
    }  \\ 
    \midrule 
     1. Place buttermilk, butter or margarine, sugar, salt, flour, whole wheat flour, and yeast into pan of bread machine.$\backslash$n2. Bake on White Bread setting.$\backslash$n3. Cool on wire racks before slicing. &
     1. Mix all ingredients together.$\backslash$n2. Let the dough rise for 1 hour.$\backslash$n3. Punch down the dough.$\backslash$n4. Let the dough rise for 1 hour.$\backslash$n5. Shape the dough into loaves.$\backslash$n6. Let the loaves rise for 1 hour.$\backslash$n7. Bake at 350 degrees for 30 minutes. \\
    
    \bottomrule
    \end{tabularx}
\end{adjustbox}
  
\end{center}
\caption{Sample recipes produced by the LLaVA-Chef-S3 model.}
  \label{tab:llavachefv3_best1}
  \Description[]{}
\end{figure*}


\begin{table*}[!ht]
    \centering
 \begin{adjustbox}{width=0.83\textwidth}
    \begin{tabular}{|l|l|c|c|c|c|c|c|c|c|c|c|c|}
\hline

Method & Inputs & BLEU-1 & BLEU-2 & BLEU-3 & BLEU-4 & SacreBLEU & METEOR & ROUGE-1 & ROUGE-2 & ROUGE-L  & CIDEr & Perplexity $\downarrow$ \\
\hline
\multirow{6}{*}{LLaVA}  & $X_i$ & 0.152 & 0.057 &0.024&0.011& 0.015 & 0.096 & 0.178 & 0.026 & 0.1 & 0.004 & 16.76 \\
 & $X_t$ & 0.213 &  0.101 &0.051&0.027 & 0.03 & 0.144 & 0.262 & 0.059 & 0.136 & 0.019 & 32.94  \\
 & $X_i$ + $X_t$ & 0.158 & 0.061 &0.025& 0.011& 0.016 & 0.104 & 0.195 & 0.029 & 0.109 & 0.005  &  14.62 \\
 & $X_i$ + $X_{ing}$ & 0.277 & 0.144 &0.079&0.045& 0.054 & 0.182 & 0.349 & 0.095 & 0.177 & 0.036 & 2.77   \\
 & $X_t$ + $X_{ing}$ & 0.293 & 0.157 &0.088& 0.051& 0.061 & 0.2 & 0.367 & 0.106 & 0.183 & 0.046 & 2.38  \\
 & $X_i$ + $X_t$ + $X_{ing}$ & 0.29 & 0.154 &0.087& 0.051 & 0.06 & 0.2 & 0.367 & 0.105 & 0.182 & 0.041 & 2.6   \\

\hline
\hline
\multirow{6}{*}{LLaVA-Chef-S1}  & $X_i$ & 0.144 & 0.059 &0.027&0.014& 0.021 & 0.067 & 0.207 & 0.035 & 0.13 & 0.014 & 3.54 \\
  & $X_t$ & 0.225 & 0.111 &0.06&0.035& 0.048 & 0.107 & 0.278 & 0.069 & 0.158 &  0.065 & 3.54   \\
  & $X_i$ + $X_t$ & 0.227 & 0.115 &0.063 & 0.037 & 0.05 & 0.108 & 0.283 & 0.073 & 0.162 &  0.065 & 2.89   \\
  & $X_i$ + $X_{ing}$ & 0.253 & 0.144 &0.088&0.055& 0.074 & 0.137 & 0.357 & 0.113 & 0.196 & 0.168 & 3.0  \\
  & $X_t$ + $X_{ing}$ & 0.325 & 0.191 &0.119& 0.076& 0.097 & 0.16 & 0.404 & 0.14 & 0.218 &  0.201 & 2.63   \\

  & $X_i$ + $X_t$ + $X_{ing}$ & 0.327 & 0.192 &0.116& 0.074& 0.096 & 0.16 & 0.405 & 0.14 & 0.219 &  0.198 &  3.54 \\
\hline

\hline
\hline
\multirow{6}{*}{LLaVA-Chef-S2}  & $X_i$ & 0.188 & 0.082 &0.04 &0.021& 0.047 & 0.078 & 0.223 & 0.042 & 0.136 & 0.016 & 2.3   \\
  & $X_t$ & 0.253 & 0.13 &0.072 &0.042 & 0.076 & 0.11 & 0.294 & 0.079 & 0.166 &  0.078 & 2.71   \\
  & $X_i$ + $X_t$ & 0.256 & 0.131 &0.072&0.042& 0.076 & 0.111 & 0.298 & 0.078 & 0.167 &  0.081 & 2.74 \\
  & $X_i$ + $X_{ing}$ & 0.308 & 0.175 &0.106& 0.067& 0.111 & 0.151 & 0.378 & 0.12 & 0.204 & 0.174 & 2.77   \\
  & $X_t$+ $X_{ing}$ & 0.338 & 0.197 &0.121&0.078& 0.124 & 0.163 & 0.407 & 0.138 & 0.219 &  0.179 & 2.6   \\

  & $X_i$+ $X_t$ + $X_{ing}$ & 0.337 & 0.196 &0.121&0.078& 0.124 & 0.163 & 0.41 & 0.14 & 0.221 &  0.189 &  2.63 \\

\hline
\hline
\multirow{6}{*}{LLaVA-Chef-S3}  & $X_i$ & 0.209 & 0.092 & 0.042 & 0.021 & 0.082 & 0.091 & 0.242 & 0.048 & 0.135 & 0.011 & \textbf{1.72}  \\
  & $X_t$ & 0.283 & 0.149 & 0.081 & 0.047 & 0.116 & 0.142 & 0.37 & 0.108 & 0.193 &  0.094 & 2.08   \\
  & $X_i$ + $X_t$ & 0.293 & 0.155 & 0.086 & 0.049 & 0.123 & 0.146 & 0.373 & 0.11 &  0.195 & 0.102 & 2.05 \\
  & $X_i$ + $X_{ing}$ & 0.337 & 0.197 & 0.12 & 0.077 & 0.156 & 0.177 & 0.45 & 0.156 & 0.232 & 0.203 & 2.43   \\
  & $X_t$ + $X_{ing}$ & 0.362 & 0.213 & 0.132 & 0.086 & 0.16 & 0.187 & 0.471 & 0.166 & \textbf{0.249} &  0.215 & 2.41   \\

   &  $X_i$ + $X_t$ + $X_{ing}$ &  \textbf{0.366} &  \textbf{0.218} & \textbf{0.137} & \textbf{0.09} &  \textbf{0.17} &  \textbf{0.189} &   \textbf{0.473} &  \textbf{0.17} &  0.24 &   \textbf{0.242} &  2.4  \\
\hline

\end{tabular}
\end{adjustbox}
\caption{
We analyzed the role of different information sources in generating cooking instructions on the test1K subset of the Recipe1M test set. 
While food images provide valuable context, our ablation study reveals that food names and ingredients are essential for accurate results.}
\label{tab:ablation1}
\end{table*}


\subsection{Ablation Study}
\paragraph{\textbf{Improvement through multi-stage training:}}
LLaVA-Chef's training in a multi-stage setup demonstrates a gradual improvement in its recipe generation capabilities, as evident from scores in Table \ref{tab:ablation1}. Pre-trained LLaVA generates recipes with hallucinations and sometimes discrepancies from the ground truth. However, LLaVA-Chef improves in every stage by a noticeable margin. The example in Figure~\ref{fig:pred_stages1} shows that LLaVA-Chef-S1 correctly estimates the required temperature, but it misjudges the mixing pattern of the ingredients and baking time. In Stage-2, it instructs to combine all ingredients in one step, though it misses an ingredient (garlic). While minor discrepancies in instructions remain, the ability to accurately list all ingredients in Stage-3 highlights the model's learning trajectory and potential.

\paragraph{\textbf{Impact of scaling loss:}} 
As discussed earlier, after stage-2, we introduce a penalty by scaling the loss based on BLEU and Rouge scores and continued training for 2 epochs. The resulting model is LLaVA-Chef-S3. To evaluate the improvement through this additional penalty, we continued training of S2 for two more epochs with only cross-entropy loss, the resulting model is LLaVA-Chef-S22. As evident in Table \ref{tab:ablation2}, although both models have been trained for 2 additional epochs after S2, the difference in performance directly reflects the impact of our novel penalty formulation.  

\begin{table}[htbp]
    \centering
 \begin{adjustbox}{width=0.44\textwidth}
    \begin{tabular}{|l|l|c|c|c|c|c|}
\hline

Method & Loss & BLEU-1 & BLEU-4 & SacreBLEU & ROUGE-L  & CIDEr  \\
\hline
LLaVA-Chef-S22  & $L_{CE}$ & \textbf{0.372}  & 0.08 & 0.158 &  0.227 &  0.191    \\
\hline
 LLaVA-Chef-S3 &  $L_{BR} \times L_{CE}$ &  0.366 &   \textbf{0.09} &  \textbf{0.17} &   \textbf{0.24} &   \textbf{0.242}  \\
\hline
\end{tabular}
\end{adjustbox}
\caption{Effect of language quality penalty loss function.}
\label{tab:ablation2}
\end{table}
\paragraph{\textbf{Impact of input attributes:}}
We also assess LLaVA and LLaVA-Chef models under various input configurations, including scenarios where only the food image is provided (excluding title and ingredients), the food image with the title (excluding ingredients), and title with ingredients. The evaluation is conducted on the $test1k$ test set, and the outcomes are summarized in Table \ref{tab:ablation1}. Our LLaVA-Chef model improves in each steps, outperforms others, showing the effectiveness of our multi-stage approach. Our initial observations revealed that images alone convey less semantic information about the food compared to food names. This is likely due to the limitations of visual information captured in images. Nevertheless, title and ingredients remain a crucial factor in recipe generation as evident by increase in scores when both are input to the model. 

Incorporating images alongside textual prompts failed to improve the performance of a pre-trained LLaVA model for recipe generation tasks. This might be attributed to limitations in the model's ability to map visual features of food images effectively into the language space. Conversely, our fine-tuned LLaVA-Chef-S1 exhibits minimal performance enhancement from image integration, regardless of its placement alongside the title or in conjunction with both title and ingredients. LLaVA-Chef-S2 exposed to a wider variety of prompts during training, demonstrates significant improvement over LLaVA when presented with solely an image. Although titles and ingredients remain essential for generating accurate cooking instructions. Our final model, LLaVA-Chef-S3, generally achieves superior scores. Interestingly, LLaVA-Chef-S3, when prompted solely with an image $(X_i)$, achieved the lowest perplexity score, but it has underwhelming performance on other metrics. Notably, while all models, including Chef-Transformer, exhibited CIDEr scores lower than 1, our final model achieved an impressive improvement of nearly 24 points in this metric. 

\section{Conclusion}
This work presents LLaVA-Chef, a multi-modal model trained for recipe generation. Through systematic evaluation of prominent open-source LLMs, we identified LLaVA as the optimal starting point. Subsequent fine-tuning utilized specially curated prompts to progressively guide the model's adaptation to the food domain. Our multi-stage method incorporated diverse prompts and a novel language quality penalty loss function, leading to significant performance gains that surpass existing methods by noticeable margins yielding state-of-the-art performance for this task. Notably, the final model, LLaVA-Chef-S3, generates recipes that are demonstrably more accurate and detailed than its predecessors, often featuring precise ingredient mentions that enhance understandability and sometimes even surpasses the quality of human-authored ground truth recipes. These findings highlight the effectiveness of our stage-wise fine-tuning approach and paves the way for further advancements for food-related tasks. While LLaVA-Chef outperforms other models in recipe generation tasks, it lacks the capability to suggest ingredient substitutions with accompanying justifications regarding health impacts. Future research will focus on expanding LLaVA-Chef's functionalities beyond recipe generation to incorporate ingredient substitution while considering dietary constraints. Another interesting direction is to consider numeric information in evaluating generated recipes, such as cooking time or temperature, ingredient quantities, and so on.



\clearpage
\bibliographystyle{ACM-Reference-Format}
\bibliography{cikm}

\end{document}